\documentclass[10pt,twocolumn,letterpaper]{article}

\usepackage{iccv}
\usepackage{times}
\usepackage{epsfig}
\usepackage{graphicx}
\usepackage{amsmath}
\usepackage{amssymb}
\usepackage{amsbsy}
\usepackage{epstopdf}
\usepackage{float}
\usepackage{xcolor}
\usepackage{tabularx, booktabs}
\usepackage{multirow}
\usepackage{commath}
\usepackage{subcaption}
\usepackage{blindtext}
\usepackage{amsmath}
\usepackage{mathtools}
\usepackage[percent]{overpic}

\DeclareMathOperator{\sigmoid}{sigmoid}

\usepackage[pagebackref=true,breaklinks=true,letterpaper=true,colorlinks,bookmarks=false]{hyperref}

\newcommand{\ignore}[1]{}
\newcolumntype{Y}{>{\centering\arraybackslash}X}
\newcolumntype{x}[1]{>{\centering\let\newline\\\arraybackslash\hspace{0pt}}p{#1}}

\iccvfinalcopy 

\def\iccvPaperID{5654} 

\ificcvfinal\fi

\begin{document}

\title{Let's See Clearly: Contaminant Artifact Removal for Moving Cameras}

\author{Xiaoyu Li$^{1}$ \qquad Bo Zhang$^{2}$ \qquad Jing Liao$^{3}$ \qquad Pedro V. Sander$^{1}$ \vspace{8pt}\\
$^{1}$The Hong Kong University of Science and Technology \qquad
$^{2}$Microsoft Research Asia \qquad \\
$^{3}$City University of Hong Kong \qquad\\
}

\twocolumn[{
\renewcommand\twocolumn[1][]{#1}

\maketitle
\ificcvfinal\thispagestyle{empty}\fi

\centering
\vspace{-0.6cm}
\includegraphics[width=\textwidth]{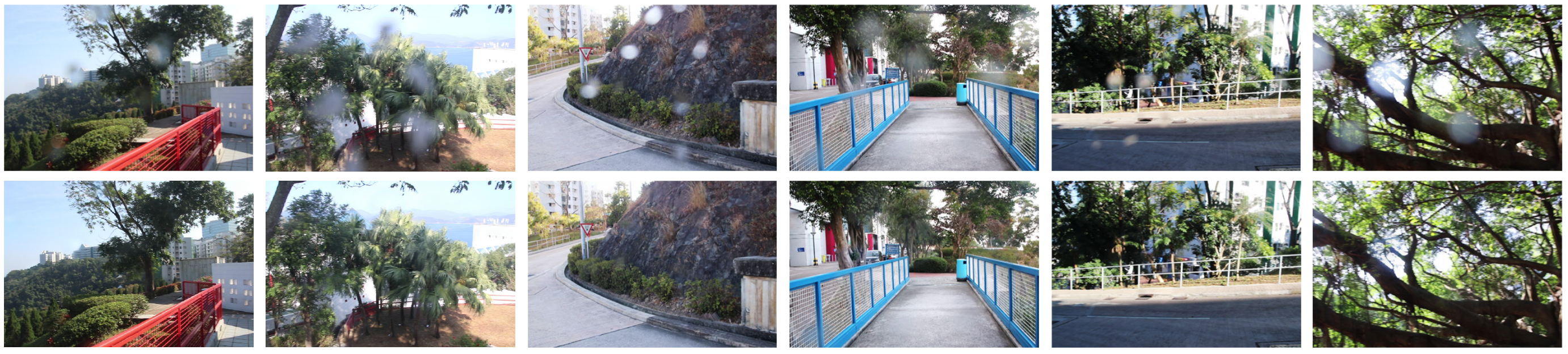}
\vspace{-0.6cm}
\captionsetup{type=figure}
\caption{Contaminant removal for video frames captured by dirty lens camera. Contaminants on the lens, \eg, dust, dirt and moisture, cause spatially variant photography artifacts (first row). Our method restores these contaminant artifacts by leveraging the spatio-temporal consistency from multiple frames (second row). Please refer to our \emph{supplementary material} for video results.}
\vspace{0.4cm}
\label{fig:teaserfigure}
}]

\begin{abstract}
    Contaminants such as dust, dirt and moisture adhering to the camera lens can greatly affect the quality and clarity of the resulting image or video. In this paper, we propose a video restoration method to automatically remove these contaminants and produce a clean video. Our approach first seeks to detect attention maps that indicate the regions that need to be restored. In order to leverage the corresponding clean pixels from adjacent frames, we propose a flow completion module to hallucinate the flow of the background scene to the attention regions degraded by the contaminants. Guided by the attention maps and completed flows, we propose a recurrent technique to restore the input frame by fetching clean pixels from adjacent frames. Finally, a multi-frame processing stage is used to further process the entire video sequence in order to enforce temporal consistency. The entire network is trained on a synthetic dataset that approximates the physical lighting properties of contaminant artifacts. This new dataset and our novel framework lead to our method that is able to address different contaminants and outperforms competitive restoration approaches both qualitatively and quantitatively.
\end{abstract}

\vspace{-0.4cm}

\section{Introduction}

As imaging devices have become ubiquitous, the ability to take photographs and videos everywhere and anytime has increased significantly. Mobile cameras, action cameras, surveillance cameras, and the sensors of autonomous driving cars are often exposed to the harsh environment in which contaminants, \eg, dust, mud and moisture, that adhere to the lens, will cause deterioration of image quality. Figure~\ref{fig:teaserfigure} shows some examples of dirty lens artifacts, where the visibility of the scene radiance is partially affected by the absorption and reflection of the contaminants along the light path~\cite{gu2009removing}. These undesired artifacts not only are aesthetically disturbing but also bring difficulty for subsequent computer vision tasks. Although one can physically clean the lens sporadically, doing this frequently is by no means a handy solution and sometimes infeasible for real-time situations.

Since the contaminants adhere to the lens surface and thereby lie out of focus, their imaging effect can be modeled by a low-frequency light modulation~\cite{gu2009removing}, \ie, the dirty lens artifacts appear diffuse and semi-transparent with the high-frequency textures of the background scene partially preserved. This makes image or video inpainting methods~\cite{chang2019free,huang2016temporally,xu2019deep,yu2019free} inadequate for our task as they completely ignore the underlying structures and the hallucinated content. Albeit visually plausible, they may deviate significantly from the real scene. Furthermore, these works assume the completion regions are prescribed by a user-given mask, whereas our task automatically identifies the degradation region, which is inferred from camera motion.

This work is more closely related to single image artifact removal for raindrops~\cite{eigen2013restoring, hao2019learning, qian2018attentive, quan2019deep}, reflection~\cite{arvanitopoulos2017single,fan2017generic,wei2019single,zhang2018single} and thin obstructions~\cite{mccloskey2010removal}. These works typically adopt learning approaches, utilizing the spatial prior of natural images to restore the spatial variant degradation. Nonetheless, the artifact removal for a single image is inherently ill-posed, and the learned spatial prior often fails to generalize to scenes with domain gaps. To solve this, multi-frame approaches~\cite{alayrac2019visual,liu2020learning,xue2015computational} decouple the occlusion and background scene by leveraging the fact that there exists motion difference between the two layers, and the pixels occluded in one frame are likely to be revealed in other frames. In particular, the recent learning-based approach~\cite{liu2020learning} achieves remarkable quality in removing unwanted reflection and obstructions. However, this method only considers a fixed number of adjacent frames as input, which should be varied depending on the magnitude of the motion and obstruction size, whereas our recurrent scheme supports an arbitrary number of adjacent frames for restoration until convergence.

In this work, we propose a learning-based framework tailored for removing the contaminant artifacts of moving cameras. To this end, we first train the network to automatically spot the contaminant artifacts which are usually prominent in the flow maps of a video with a moving camera. As opposed to layer decomposition, we only focus on the background motion, of which the degraded region by the contaminants is hallucinated and softly blended by our flow completion network, depending on how much of the background is occluded.

In order to leverage information spanning an arbitrary number of frames, the restoration for each frame is recurrent. That is, to restore one frame, we recurrently feed the adjacent frames one by one. Guided by the completed background flow, the pixels within the artifact region can be progressively restored by referring to the corresponding clean pixels from other frames. So far the restoration operates on each input frame individually,  utilizing only the information of their adjacent frames. To produce the temporally consistent result for the whole video, we propose another multi-frame processing stage, in which we follow the same pipeline again but this time using the restored results from the last recurrent stage as input.

We train the entire framework in a supervised fashion. To achieve this, we propose a synthetic dataset that follows the imaging physics of contaminant artifacts. Extensive experiments prove that the proposed model can generalize to real dirty lens videos (as shown in Figure~\ref{fig:teaserfigure}), outperforming strong baselines both qualitatively and quantitatively. Our contributions can be summarized as follows:

\begin{itemize}
    \item We propose the first deep learning approach to specifically address the contaminant artifacts for moving cameras. The proposed method performs better than general restoration methods on real videos.
    \item A physics-inspired synthetic dataset is proposed to mimic real contaminant artifacts. 
    \item We propose a flow completion module to effectively hallucinate the background motion given the partially visible structure clue within the degraded region.
    \item The proposed recurrent scheme not only helps leverage the multiple adjacent frame information to restore individual frames but can also be reused to refine such frame-wise output and ultimately yield temporally coherent video results.
\end{itemize}

\section{Related Work}

\paragraph{Camera artifact removal.} The pioneer work~\cite{gu2009removing} proposes a physics-based method to remove the dirty lens artifact, yet the point-wise restoration they propose cannot handle complex artifacts. 
Following works, on the other hand, merely focus on the contaminant detection~\cite{akkala2016lens, chen2016dust, uricar2021let, yang2017dust}, but do not study how to give a clean image with contaminant removal. Indeed, the artifacts region can be restored with a follow-up content completion~\cite{mccloskey2010removal,liu2018image,yu2019free,nazeri2019edgeconnect}, yet this will totally neglect the underlying structure within the degraded region. In comparison, we jointly consider the artifact localization and restoration in a single framework that utilizes the partially visible structures as much as possible. Notably, lens flare or glare is another common lens artifact that plagues the photography in which the scene is also partially obstructed. Nonetheless, existing solutions~\cite{asha2019auto,raskar2008glare,vitoria2019automatic} focus on single image restoration which is inherently ill-posed, whereas our method explicitly utilizes multi-frame information captured by moving cameras.

\vspace{-0.4cm}

\paragraph{Adherent raindrop removal.} A number of methods have been proposed to address raindrops attached to glass windows, windscreens, \etal, mostly for single image~\cite{eigen2013restoring, qian2018attentive, hao2019learning, quan2019deep}. Several methods have been proposed to remove raindrops from video~\cite{roser2009video,yamashita2009noises, you2013adherent, you2014raindrop, you2015adherent}. However, after detecting the raindrops using the spatial-temporal information, these methods rely on off-the-shelf video inpainting techniques to restore the video, which does not fully utilize the partially visible details within the raindrops. Besides, both the raindrop detection and restoration are optimized separately. Recently, Liu~\etal~\cite{liu2020learning} present a learning approach for removing unwanted obstructions which also include semi-transparent raindrops. Instead of formulating the problem as layer decomposition, we only consider the scene motion and use a recurrent scheme to consider information from an arbitrary number of adjacent frames. Besides, our method does not require time-consuming online optimization as post-processing for handling real-world sequences.

\vspace{-0.4cm}

\paragraph{Video-based restoration.} Video-based restoration such as video inpainting, video denoising and video deblur utilizes spatial-temporal information for restoration. One typical application is rain streak removal. While prior approaches rely on hand-crafted features~\cite{barnum2010analysis,garg2004detection,garg2005does,garg2006photorealistic,garg2007vision,santhaseelan2015utilizing,zhang2006rain}, 
recent prevalent methods resort to deep neural networks~\cite{chen2018robust,li2018video}. Although some of them also employ recurrent scheme~\cite{liu2018d3r, liu2018erase, yang2019frame} to leverage temporal information, additional modules like flow completion, multi-frame processing have been uniquely considered for our problem. 


\section{Method}

\begin{figure}
	\centering
	\begin{overpic}
    [scale=0.335]{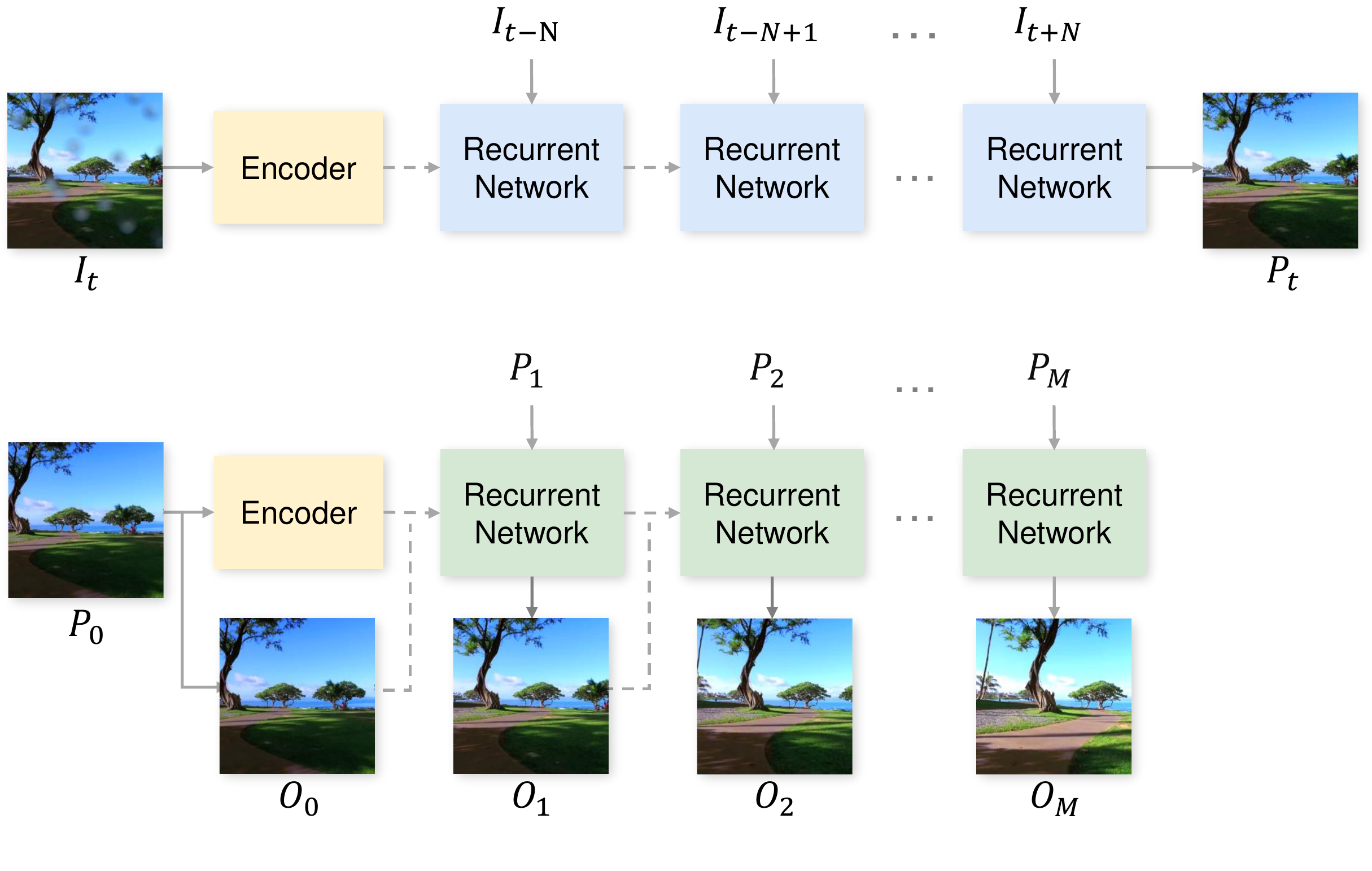}
    \put(29,41){\footnotesize (a) Single-frame restoration}
    \put(29,0){\footnotesize (b) Multi-frame processing}
    \end{overpic}
	\vspace{-0.4cm}
	\caption[width=\linewidth]{Overview of our two-stage recurrent network for contaminant removal. (a) In the single-frame restoration stage, frame $I_t$ is progressively restored by feeding its adjacent frames one by one. (b) The multi-frame processing stage takes the frame-wise results $\{P_t\}$ as input and recurrently processes them to produce a temporal coherent result $\{O_t\}$.}
	\label{fig:overview_0}
	\vspace{-0.4cm}
\end{figure}

Figure~\ref{fig:overview_0} illustrates the proposed two-stage recurrent framework. Given an input frame $I_t$ suffering from contaminant artifacts, we first gradually restore the degraded region by iteratively utilizing the adjacent frames $I_{t-N\leq k \leq t+N}$ that may reveal some new clean pixels under the camera motion. This is achieved by aligning the frames with the hallucinated flow. This way, we obtain frame-wise intermediate outputs $\{P_t\}$, which are further fed into the multi-frame processing stage and yield the frames $\{O_t\}$ that consider the temporal consistency relative to the outputs at an earlier time. Next, we introduce a synthetic dataset that realistically emulates the contaminant artifacts for training (Section~\ref{sec:dataset}). Then we elaborate on the details of the single-frame restoration (Section~\ref{sec:singlefrm}) and the multi-frame processing (Section~\ref{sec:multifrm}), respectively.

\subsection{Dataset Construction}
\label{sec:dataset}

\begin{figure}
	\centering
	\includegraphics[width=\linewidth]{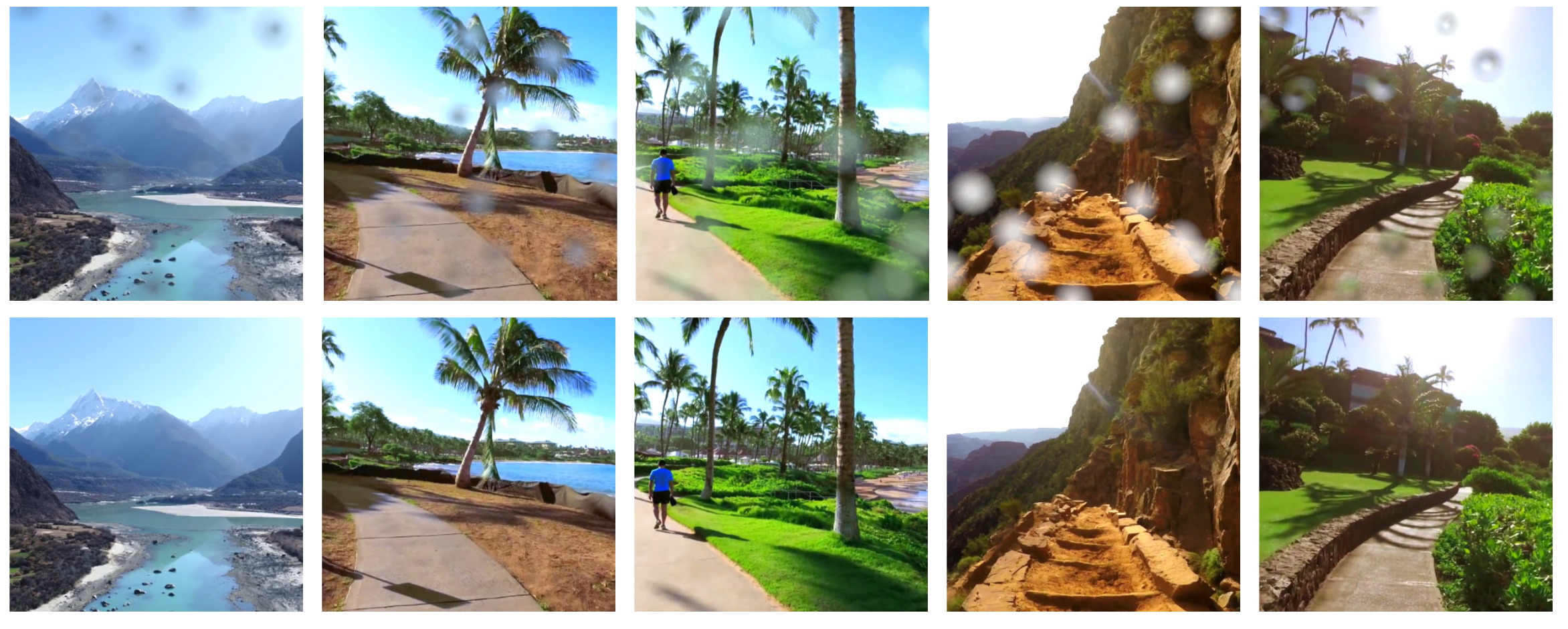}
	\vspace{-0.6cm}
	\caption[width=\linewidth]{Samples from our synthetic dataset. The first row shows images with contaminant artifacts and the second row shows the corresponding ground truth images.}
	\label{fig:dataset}
	\vspace{-0.4cm}
\end{figure}

It is challenging to obtain large quantities of well-aligned video pairs of real scenes, so we synthesize a training dataset that covers realistic and diverse contaminant artifacts. To this end, we render images following the physics model~\cite{gu2009removing} about how the contaminants affect the image irradiance. Specifically, we use Blender for the rendering. We collect a large number of moving camera videos as source frames, which serve as the scene textures representing our scene. Between this background scene and the camera, we put a glass layer that is fully transparent (with the index of refraction set to $1$) to model a contaminant-free camera lens. We model the contaminants with manually deformed particles, whose material is a mixture of different shaders: the glass shader adds some refraction, the emission shader contributes some radiance so as to emulate the scattering due to the lens dirt, and the transparent shader models the light attenuation caused by the contaminants. By stochastically varying the parameters of these shaders, we are able to simulate the effect of common contaminant materials. Figure~\ref{fig:dataset} shows examples of our rendered images. The synthetic samples closely mimic the real contaminated images and have a large variation to cover common situations in real photos.

\begin{figure*}
	\centering
	\begin{overpic}
    [scale=0.52]{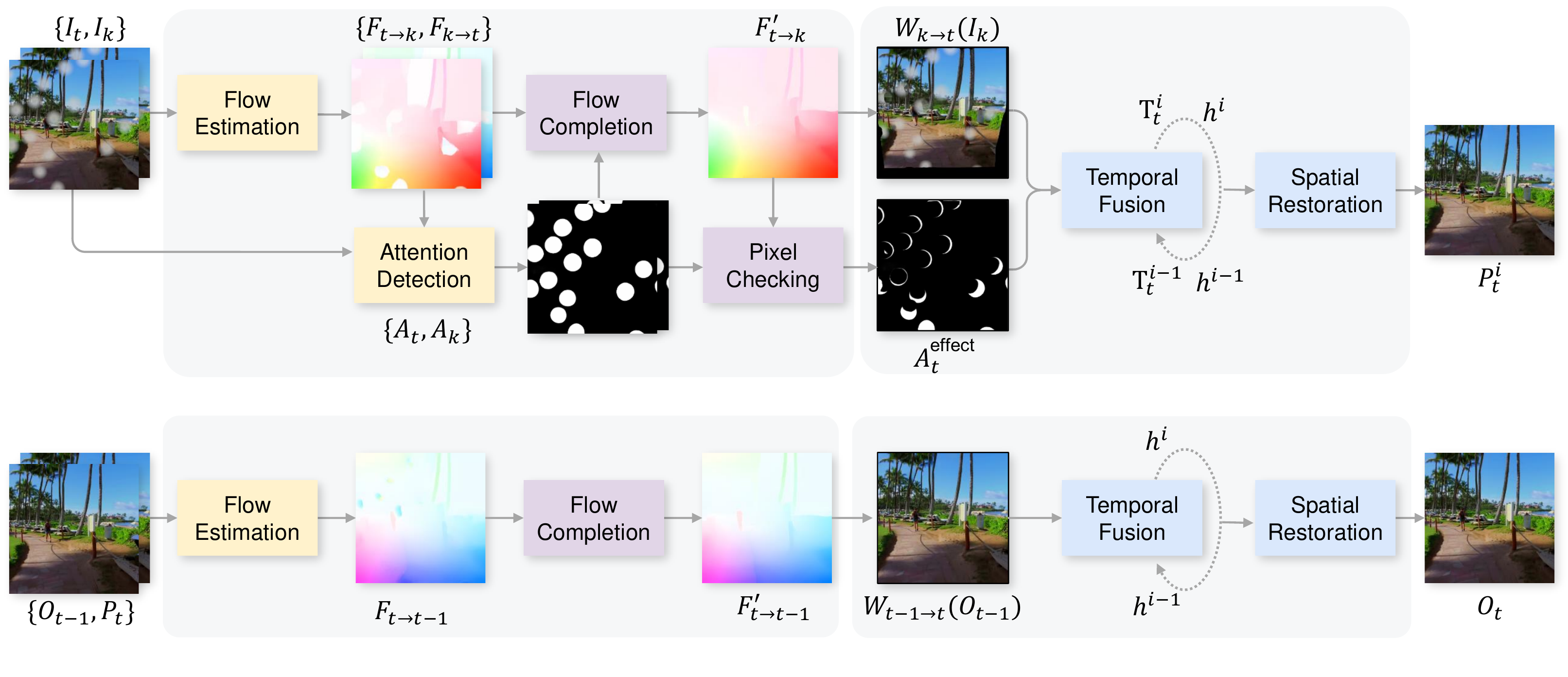}
    \put(40,17.6){\footnotesize (a) Single-frame restoration}
    \put(40,0.9){\footnotesize (b) Multi-frame processing}
    \end{overpic}
	\vspace{-0.2cm}
	\caption[width=\linewidth]{Overview of the recurrent pipeline for (a) single-frame restoration and (b) multi-frame processing.}
	\label{fig:overview_1}
	\vspace{-0.4cm}
\end{figure*}

\subsection{Single-frame Restoration}

\label{sec:singlefrm}
In this stage, we aim to remove the artifacts from frame $I_t$ by recurrently referring to the adjacent frames $\{I_k\}$. The procedures for the single-frame restoration are depicted in Figure~\ref{fig:overview_1}~(a). We first estimate the bidirectional flows $\{F_{t\rightarrow k}, F_{k\rightarrow t}\}$ between the two frames $\{I_t, I_k\}$ and detect the attention maps $\{A_t, A_k\}$ to localize the degraded region based on the flows. Guided by the attention maps, we complete the background motion by a flow completion module so that we can warp the reference frame towards the input accordingly. The pixel checking module validates whether the pixel in the warped reference $\mathcal{W}_{k\rightarrow t}(I_k)$ can be used to restore the corresponding contaminated pixels in $I_t$. Next, a recurrent temporal fusion module updates the restored result from $T_t^{i-1}$ to $T_t^{i}$ by leveraging the effective clean pixels from $\mathcal{W}_{k\rightarrow t}(I_k)$ as well as the recurrent hidden state $h^{i-1}$ that comes from the last iteration $(i-1)$. Finally, the hidden representation is decoded to the image output $P_t^{i}$ with a spatial restoration module.

\vspace{-0.4cm}

\paragraph{Flow estimation \& attention detection.}
We first estimate the optical flows $\{F_{t\rightarrow k}, F_{k\rightarrow t}\}$ between the input $I_t$ and its adjacent frame $I_k$ using the off-the-shelf RAFT model~\cite{teed2020raft}. As shown in Figure~\ref{fig:overview_1}~(a), the contaminants become prominent in the estimated flow. Therefore we could utilize it to help predict the attention map that indicates the degraded region. Specifically, we adopt an U-Net~\cite{ronneberger2015u} to estimate the attention map $A_t$ for $I_t$ using the information of flow $F_{t\rightarrow k}$ assisted with the frame $I_t$. The network is trained with a binary cross entropy (BCE) loss between $A_t$ and the ground truth $A_t^{\rm gt}$:
\vspace{-0.1cm}
\begin{equation}
    \mathcal{L}_{\rm att} = - \frac{1}{HW} \sum_{p} A_t^{\rm gt} \log A_t + (1 - A_t^{\rm gt}) \log (1 - A_t)
\end{equation}
where $p$ indexes the pixels and $HW$ is the image resolution. Similarly, the attention map $A_k$ of $I_k$ can be estimated using the inverse flow $F_{k \rightarrow t}$ and frame $I_k$. Here, a higher value for $A_t$ indicates a higher possibility of being occluded by the contaminants.

\vspace{-0.4cm}

\paragraph{Flow completion \& pixel checking.}

\begin{figure}
    \footnotesize
	\centering
	\includegraphics[width=0.98\linewidth]{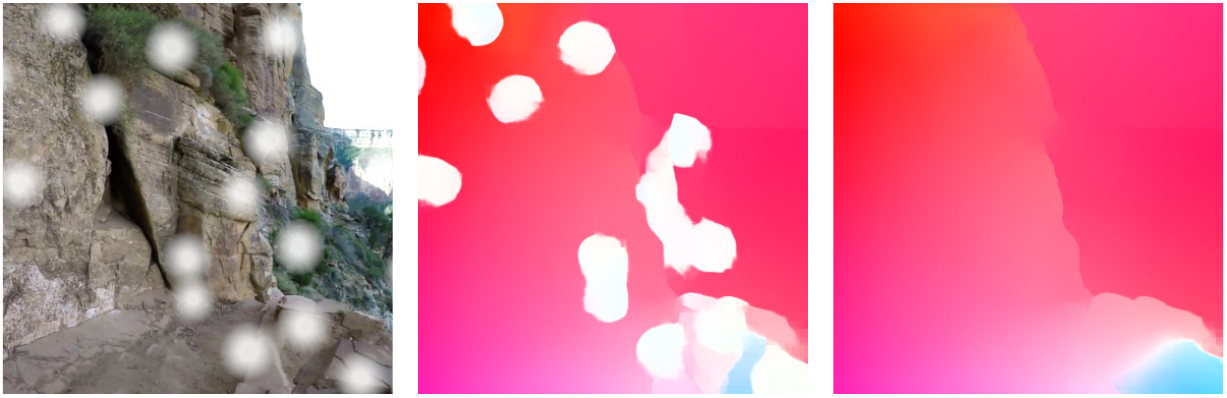}
	\begin{tabularx}{0.98\linewidth}{YYY}
        $I_k$ & (a) Degraded Flow & (b) Completed Flow \\
    \end{tabularx}
	\includegraphics[width=0.98\linewidth]{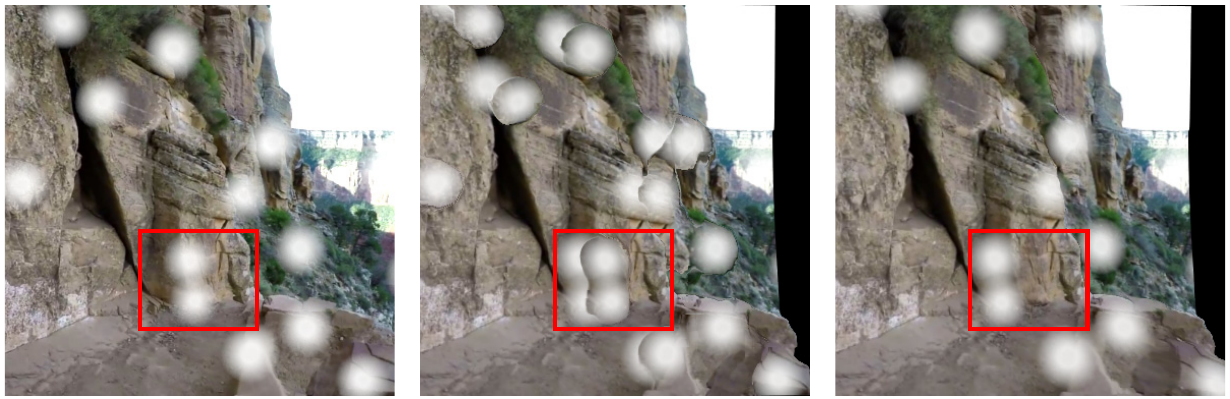}
	\begin{tabularx}{0.98\linewidth}{YYY}
        $I_t$ & Warped $I_k$ by (a) & Warped $I_k$ by (b) \\
    \end{tabularx}
    \vspace{-0.2cm}
	\caption[width=\linewidth]{The effect of our flow completion module. The middle of the marked region is still degraded with the warping $\mathcal{W}_{k\to t}(I_k)$ using the degraded flow but corrected filled in with the completed flow.}
	\vspace{-0.3cm}
	\label{fig:flow}
\end{figure}

Due to the motion difference between the fast-moving background and relatively static contaminants, the pixels degraded in one frame could be revealed in adjacent frames. Hence, we could utilize this fact to restore the video frames. In order to leverage the corresponding clean pixels from the reference frames, we need to hallucinate the flow of the background scene that is unreliably estimated within the degraded region. To this end, we propose a flow completion module whose effect is shown in Figure~\ref{fig:flow}: the pixels within the degraded region can only be correctly filled with clean pixels according to the completed flow.  

Notably, the estimated flow within the degraded region may not be unreliable all the time in that the flow module may leverage partially visible structures and induce correct flow estimation. As such, the flow completion should flexibly hallucinate the flow depending on how much the background structures are visible. Therefore, we propose a feature fusion layer that dynamically fuses the features of two branches: the input and the flow hallucinated from the scratch, according to a fusion weight map $\alpha$:
\begin{equation}
	f_{\rm out} = {f}_{\rm in}  \odot \alpha + \mathcal{G}_l (f_{\rm in}) \odot (1 - \alpha)
\end{equation}
where $f_{\rm in}$ and ${f}_{\rm out}$ are the input and output features respectively, $\mathcal{G}_l$ denotes the mapping function, and $\odot$ is the Hadamard operator. The fusion weight map is obtained from the input feature map with layers $\mathcal{G}_{\alpha}$ followed by a sigmoid squashing function:
\begin{equation}
    \alpha = \sigmoid (\mathcal{G}_{\alpha}(f_{\rm in}))
\end{equation}
Larger values in $\alpha$ denote that the degraded pixel is more visible, so the flow completion is more confident to the flow input and resorts less to the hallucinated values. 

The flow completion module has an autoencoder architecture whose encoder consists of six such fusion layers, whereas we place four fusion layers with dilation at the bottleneck. The decoder, on the other hand, adopts the upsampling module as~\cite{teed2020raft}, \ie, learning a weighted combination of a local $3 \times 3$ neighbors at coarse resolution, which we find beneficial to produce a sharper flow estimation compared with the traditional bilinear upsampling. We enforce the flow completion with the $\mathcal{L}_1$ loss between the output $F^{\prime}_{t \rightarrow k}$ and the ground truth $F^{\rm gt}_{t \rightarrow k}$,
\begin{equation}
	\mathcal{L}_{\rm multi} = \norm{ F^{\prime}_{t \rightarrow k} - F^{\rm gt}_{t \rightarrow k}}_1.
\end{equation}

Having localized the artifact in both frames and obtained the background flow, we can determine which pixel of $I_k$ is useful to restore the degraded pixels for the current frame. We identify these effective pixels in the $\mathcal{W}_{k \to t}(I_k)$ by computing the following map:
\begin{equation}
	A^{\rm effect}_{t} = (1 - \mathcal{W}_{k \rightarrow t}(A_k)) \odot A_t,
\end{equation}
which we use to guide the following restoration modules.

\vspace{-0.4cm}

\paragraph{Spatio-temporal restoration.}

Prior approaches~\cite{you2013adherent, you2015adherent, liu2020learning} exploit the spatio-temporal information from a fixed number of frames, yet the number of adjacent frames needed for the restoration may vary depending on the magnitude of camera motion and the size of degraded region. In view of this, we propose a recurrent restoration network that provides the flexibility of feeding a varying number of adjacent frames and can thereby leverage long-term temporal information when necessary.

The whole recurrent restoration module consists of two steps: temporal fusion and spatial restoration. The temporal fusion iteratively estimates a sequence of temporarily restored results. In each iteration, the recurrent module produces an intermediate image restoration $T^{i}_t$ and a hidden state $h^i$ based on the $T^{i-1}_t$ and $h^{i-1}$ of the last iteration. We regard $I_t$ to be the initial restoration result, \ie, $T_t^0 = I_t$. The recurrent module adopts a convolutional gated recurrent unit (ConvGRU)~\cite{cho2014properties}, and the iteration process can be formulated as follows, 
\begin{equation}
\begin{aligned}
	z_i &= \sigma ({\text{Conv}}[h^{i-1}, x_i]) \\ 
	r_i &= \sigma ({\text{Conv}} [h^{i-1}, x_i]) \\
	h^{\prime} &= \tanh({\text{Conv}}[r_i \odot h^{i-1}, x_i]) \\
	h^i &= (1 - z_i) \odot h^{i-1} + z_i \odot h^{\prime}\\
\end{aligned}
\end{equation}
where $z_i$ and $r_i$ are update gate and reset gate respectively, and $x_i$ is the feature of the input which is a concatenation of the frame $I_t$, the attention map $A_t$, the warped frame $\mathcal{W}_{k \to t}(I_k)$ and the effective restoration map $A^{\rm effect}_t$. Once the hidden state is updated by the GRU block, it will pass through three convolutional layers followed by a sigmoid function to predict a blending mask $M$, which is used to attentively fuse the warping $\mathcal{W}_{k \rightarrow t}(I_k)$ and the intermediate prediction $T^{i-1}_t$:
\begin{equation}
    T^i_t = M \odot \mathcal{W}_{k \rightarrow t}(I_k) + (1 - M) \odot T^{i-1}_t
\end{equation}
We enforce such intermediate result by minimizing its mean square error against the ground truth $C_{t}$. Note that we compute the loss for all the iterations $2N$ and each iteration is accounted by different factors:
\begin{equation}
	\mathcal{L}_{\rm fusion} =  \frac{1}{2N} \sum^{2N}_{i=1} \gamma^{\left|2N-i\right|}\norm{T^{i}_{t} - C_{t}}_2^2.
\end{equation}
In the experiments we empirically use $\gamma = 0.8$.

\begin{figure*}[tb]
    \footnotesize
	\centering
	\includegraphics[width=\linewidth]{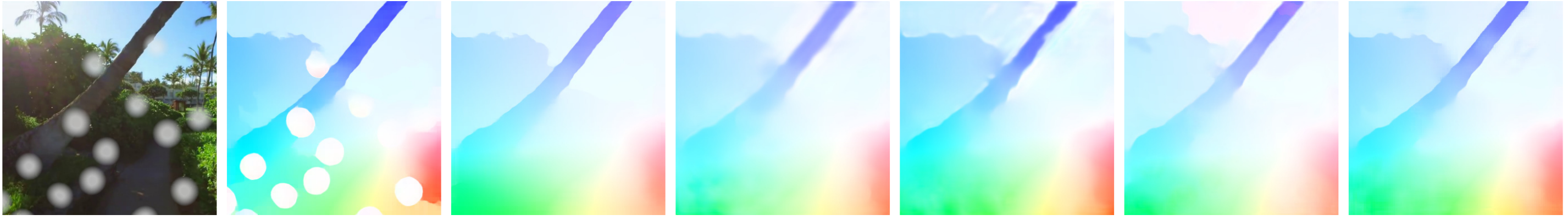}
	\begin{tabularx}{\linewidth}{YYYYYYY}
        Frame & Input Flow & GT Flow & Conv & GatedConv & FeatFusion & Ours \\
    \end{tabularx}
    \vspace{-0.4cm}
	\caption[width=\textwidth]{Example results of different network architectures for flow completion. Our full model with feature fusion layers and the upsampling module can produce a more accurate result with sharper motion boundaries.}
	\label{fig:ablation_flow}
\end{figure*}

\begin{figure*}[t]
    \footnotesize
	\centering
	\includegraphics[width=\linewidth]{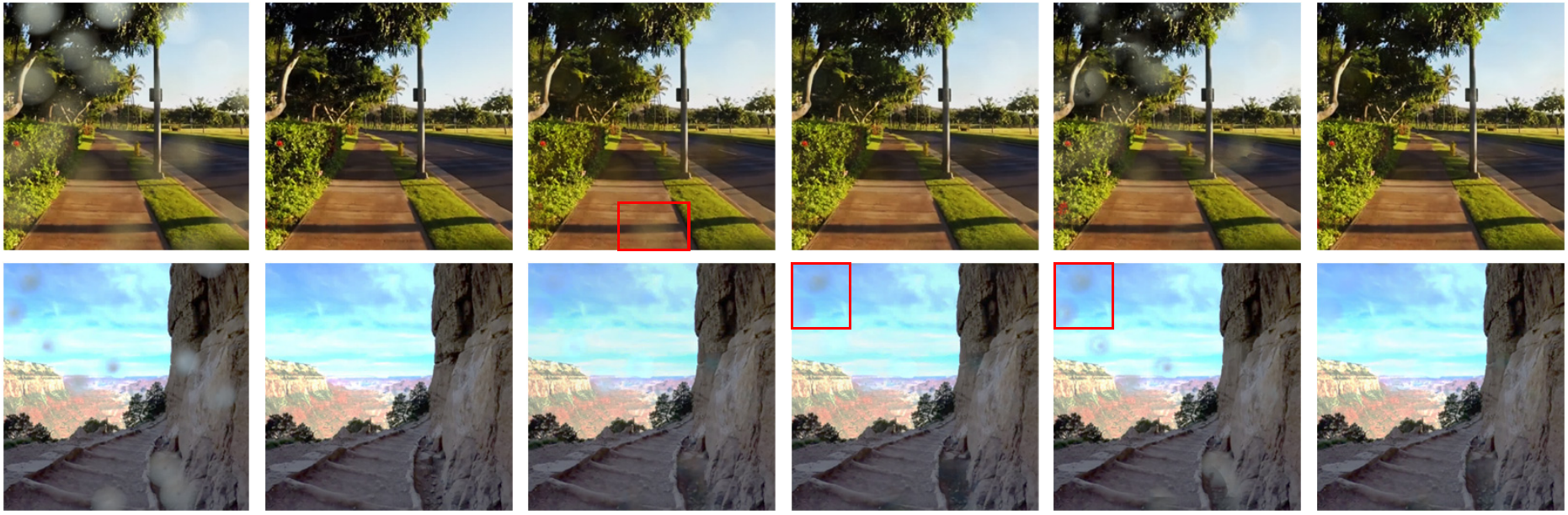}
	\begin{tabularx}{\linewidth}{YYYYYY}
        Input & GT & w/o Attention Map & w/o Flow Completion & w/o Spatial Restor. & Full Model\\
    \end{tabularx}
    \vspace{-0.4cm}
	\caption[width=\textwidth]{Ablation study of single-frame stage. Our full model can generate the results with fewer visible artifacts.}
	\vspace{-0.4cm}
	\label{fig:ablation_image}
\end{figure*}

As more adjacent frame are utilized, the restoration progressively improves. Nonetheless, there may exist scene locations occluded in all the frames, so it still requires to leverage the spatial prior for restoration. We use the contextual autoencoder architecture from~\cite{qian2018attentive} for this spatial restoration task, as shown in Figure~\ref{fig:overview_1}~(a). The network receives the temporal fusion result $T_t^i$ and the hidden state $h^i$ as the input, and learns the spatial restoration by minimizing the perceptual loss~\cite{johnson2016perceptual}:
\begin{equation}
	\mathcal{L}_{\rm spatial} = \frac{1}{2N} \frac{1}{L} \sum^{2N}_{i=1} \sum^{L}_{l=1} \norm{\phi^l({P}^{i}_{t}) - \phi^l(C_{t})}_2^2
\end{equation}
where $P_t^i$ denotes the spatial restoration output at $i$th iteration and $\phi^l$ is the $l$th layer of a pretrained VGG model~\cite{simonyan2014very}. The spatial restoration module is capable to deal with different levels of degradation during training, complementing the restoration ability of the recurrent fusion. 

In summary, we train the entire single-frame restoration network using the following objective function: 
\begin{equation}
	\mathcal{L}_{\rm single} = \mathcal{L}_{\rm att} + \mathcal{L}_{\rm multi} + \lambda_1 \mathcal{L}_{\rm fusion} + \lambda_2 \mathcal{L}_{\rm spatial}
\end{equation}
where the coefficients $\lambda_1$ and $\lambda_2$ balance different terms. In the experiments we set $\lambda_1=100$ and $\lambda_2=10$.

\subsection{Multi-frame Processing}
\label{sec:multifrm}
By far, the video frames are processed individually, based on the adjacent frames. Hence, the single frame restoration (denoted by $\Psi$) can be formulated as,
\begin{equation}
    P_t = \Psi(I_t|\{I_k\}), k \in [t-N, t+N].
\end{equation}
However, the temporal consistency over the entire output sequence cannot be guaranteed due to the nature of frame-by-frame processing. To address this, we propose a multi-frame processing stage as shown in Figure~\ref{fig:overview_0}~(b) and Figure~\ref{fig:overview_1}~(b). As opposed to the first stage that keeps refining one frame in different iterations, the multi-frame processing refines different input frames during iterations. Concretely, we feed into the outputs from the first stage in sequence, and let the network adjust $P_t$ based on the earlier output frame $O_{t-1}$, so the processing becomes:
\begin{equation}
    O_t = \Psi(P_t|O_{t-1}), \quad O_0 = P_0.
\end{equation}
A slight difference of the pipeline is that the attention detection module is no longer needed since the input frames $\{P_t\}$ are already cleaned by the frame-wise processing. Besides, we introduce a temporal loss $\mathcal{L}_{\rm temporal}$ to enforce the temporal consistency between successive outputs:
\begin{equation}
\begin{split}
    \thinmuskip=1mu
    \thickmuskip=1mu
    \mathcal{L}_{\rm temporal}  = \frac{1}{M-1} \sum_{t=2}^{M}  \Big\{e^{{-\norm{ C_{t} - \mathcal{W}_{t-1\to t}(C_{t-1})}^2_2}\big/{\mu}} \\  \times \norm{ O_{t} - \mathcal{W}_{t-1\to t}(O_{t-1}) }_1\Big\}
\end{split}
\end{equation}
where $\mathcal{W}_{t-1 \to t}(C_{t-1})$ and $\mathcal{W}_{t-1 \to t}(O_{t-1})$ are the frame warpings using the ground truth flow, $M$ is the length of the video sequence, and we set the exponential coefficient $\mu=0.02$. The overall loss for training the multi-frame processing is defined as,
\begin{equation}    \thinmuskip=0mu
    \thickmuskip=1mu
	\mathcal{L}_{\rm multi} = \mathcal{L}_{\rm multi} + \lambda_1 \mathcal{L}_{\rm fusion} + \lambda_2 \mathcal{L}_{\rm spatial} + \lambda_3 \mathcal{L}_{\rm temporal},
\end{equation}
where the newly introduced weight  $\lambda_3$ is set by $10$. 
\section{Experiments}

\subsection{Implementation}

\paragraph{Training details.}
We adopt Adam optimizer~\cite{kingma2014adam} with $\beta_1 = 0.9 $, $\beta_2 = 0.999$, a learning rate of 0.0001, and batch size of 8 images for training. Each image uses five neighboring frames as input, where the middle frame is the one to be restored in the single-frame stage. During training, We randomly crop these images from $384 \times 384$ to $256 \times 256$ for data augmentation. It first takes 300 epochs to train the single-frame stage. After that, we run the trained single-frame model on the entire dataset to generate the training set for the multi-frame stage, which takes another 50 epochs to converge. Our method is implemented using Pytorch~\cite{paszke2019pytorch}. The entire training takes approximately five days on 8x GeForce RTX 2080Ti GPUs.

\vspace{-0.4cm}

\paragraph{Datasets.} 
We render 600 video clip pairs as our training set, where each clip has 30 frames at 6fps and a resolution of $384 \times 384$. For the test set, we produce another 30 clip pairs with random rendering parameters to differentiate from the training set. We will use this test set for the quantitative evaluation in ablation studies and comparisons since the ground truth videos are available. For qualitative results, we use a Canon EOS 80D camera to capture the real videos with different contaminants adhered to the lens.

\begin{figure*}[t]
    \footnotesize
	\centering
	\includegraphics[width=\linewidth]{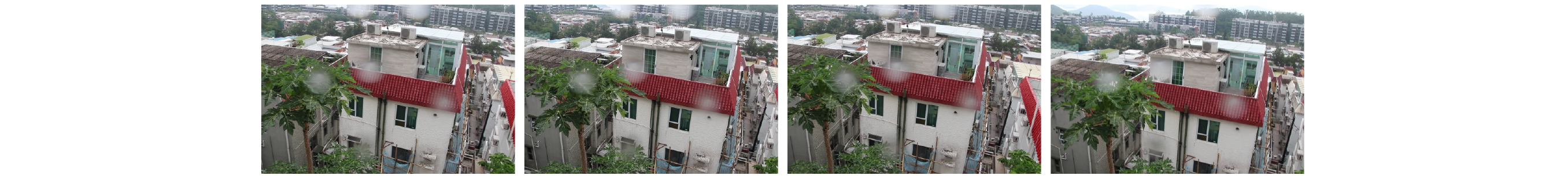}
	\begin{tabularx}{\linewidth}{YYYYYY}
         & $I_{t-1}$ & $I_{t+1}$ & $I_{t-2}$ & $I_{t+2}$ &  \\
    \end{tabularx}
    \includegraphics[width=\linewidth]{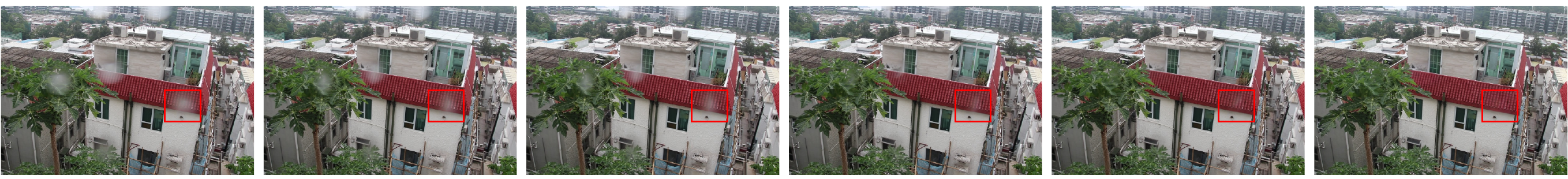}
	\begin{tabularx}{\linewidth}{YYYYYY}
         $I_{t}$ & $T^1_{t}$ & $T^2_{t}$ & $T^3_{t}$ & $T^4_{t}$ & $P^4_{t}$ \\
    \end{tabularx}
    \vspace{-0.4cm}
	\caption[width=\textwidth]{Example results showing the progressive restoration of our recurrent network. The first row is the input neighboring frames, and the second row is the corresponding restoration results.}
	\vspace{-0.2cm}
	\label{fig:res_recurrent}
\end{figure*}

\subsection{Ablation Study}

\begin{table}[t]
\centering
\footnotesize
\begin{tabular}{@{}lccccc@{}}
	\toprule
      & Input & Conv & GatedConv & FeatFusion & Ours \\
	\midrule
	EPE $\downarrow$  & 3.17 & 2.10 & 1.89 & 1.72 & \textbf{1.60} \\
	\bottomrule
\end{tabular}
\vspace{-0.2cm}
\caption{Ablation study for the flow completion network.}
\label{tab:tab_flow}
\end{table}

\paragraph{Flow completion network.} We first conduct an ablation study to demonstrate the effectiveness of our flow completion network. Different architectures with the same attention detection module are used to learn the completed flows. Specifically, we adopt the same encoder-decoder architecture with plain convolution layer (Conv), gated convolution layer (GatedConv)~\cite{yu2019free}  for image inpainting and our feature fusion layer (FeatFusion). Finally, we use the feature fusion layer and replace the decoder with our upsampling module which is our full model for flow completion task (Ours). As shown in Table~\ref{tab:tab_flow} and Figure~\ref{fig:ablation_flow}, our full model achieves the most accurate flow with the lowest endpoint error (EPE) and sharpest motion boundaries.

\vspace{-0.4cm}
\paragraph{Effectiveness of the main component.}

\begin{table}[t]
\centering
\footnotesize
\begin{tabular}{@{}lccc@{}}
	\toprule
    Model & PSNR $\uparrow$ & SSIM $\uparrow$& $E_{warp}$ $\downarrow$ \\
    \midrule
	w/o Attention Map           & 33.70 & 0.976 & 0.0046\\
	w/o Flow Completion        & 34.09 & 0.975 & 0.0046\\
	w/o Spatial Restoration     & 29.61 & 0.953 & 0.0049\\
	\cmidrule(lr){1-4}
	Full Model & \textbf{35.37} & \textbf{0.980} & \textbf{0.0045} \\
	\bottomrule
\end{tabular}
\vspace{-0.2cm}
\caption{Ablation study for the single-frame stage.}
\label{tab:tab_single}
\vspace{-0.2cm}
\end{table}

\begin{figure}[t]
    \footnotesize
	\centering
	\includegraphics[width=\linewidth]{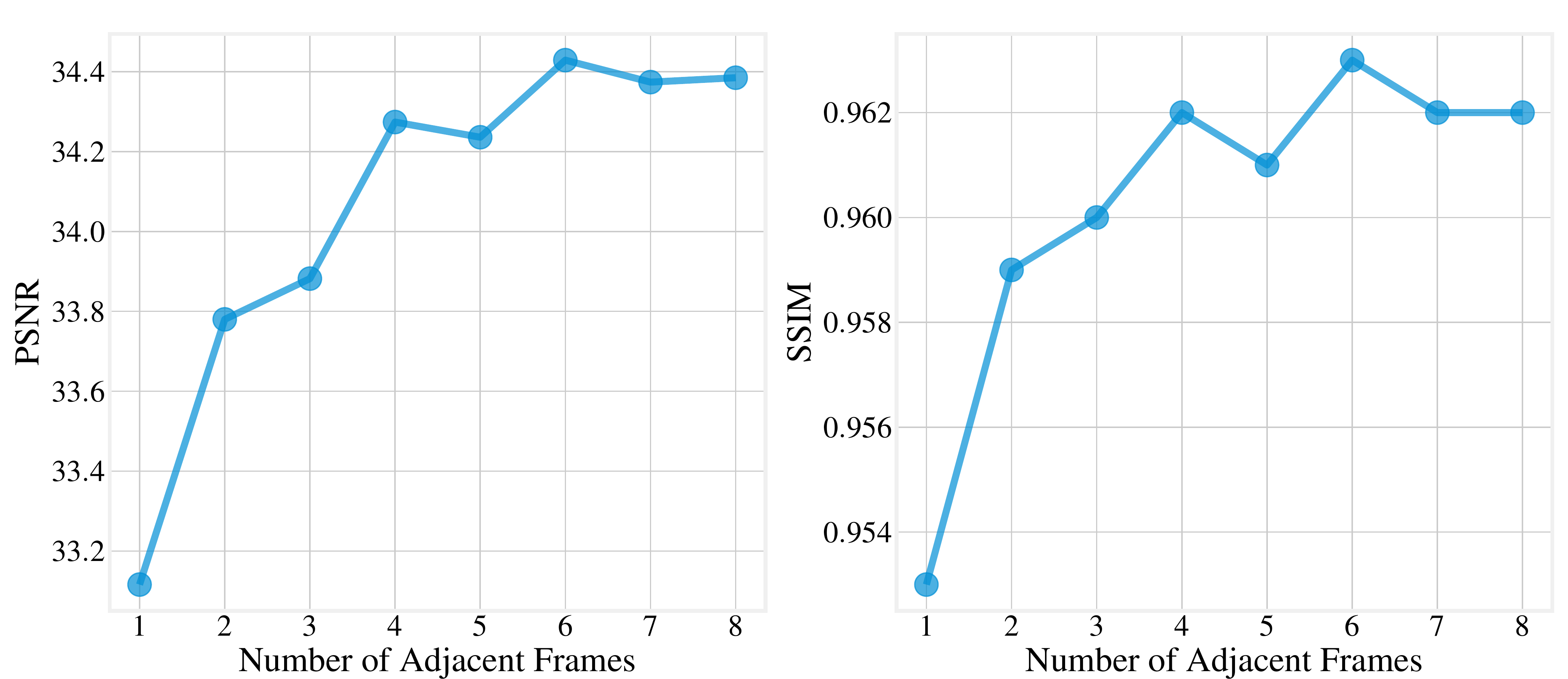}
    \vspace{-0.4cm}
	\caption[width=\textwidth]{Quality comparison of our approach with different input frames. Our method is able to use more frames for better performance until convergence.}
	\vspace{-0.4cm}
	\label{fig:res_frmnum}
\end{figure}

\begin{figure*}[t]
    \footnotesize
	\centering
	\includegraphics[width=\linewidth]{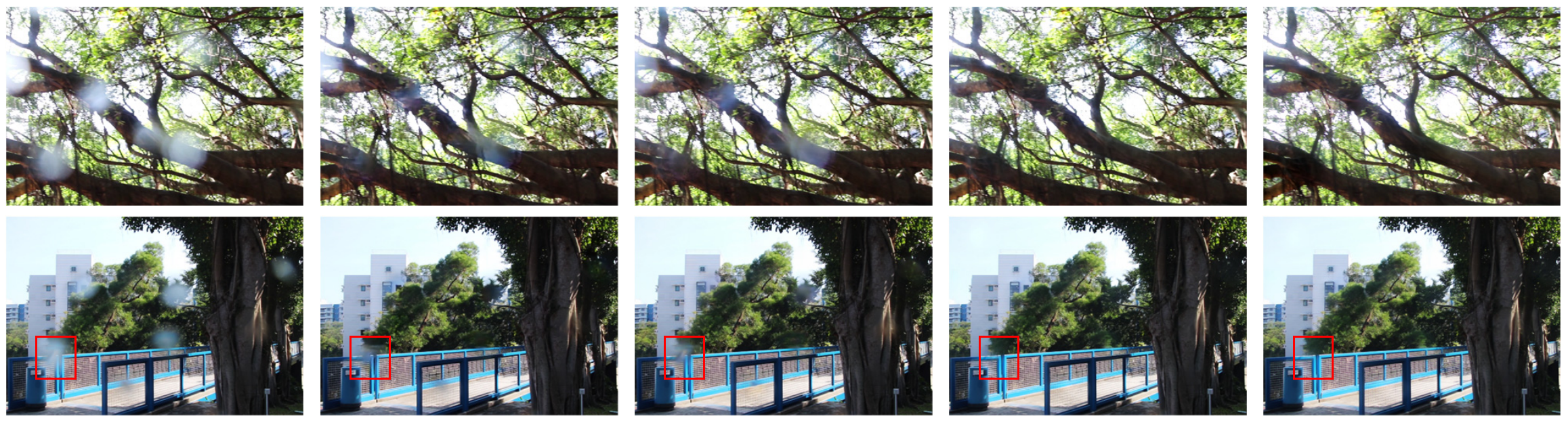}
	\begin{tabularx}{0.98\linewidth}{YYYYY}
        Input & PReNet~\cite{ren2019progressive} & AttGAN~\cite{qian2018attentive} & Ours (stage one) & Ours \\
    \end{tabularx}
    \vspace{-0.2cm}
	\caption[width=\textwidth]{Qualitative comparison with single-image restoration methods on the real images.}
	\vspace{-0.2cm}
	\label{fig:comparison_single}
\end{figure*}

\begin{figure*}[t]
    \footnotesize
	\centering
	\includegraphics[width=\linewidth]{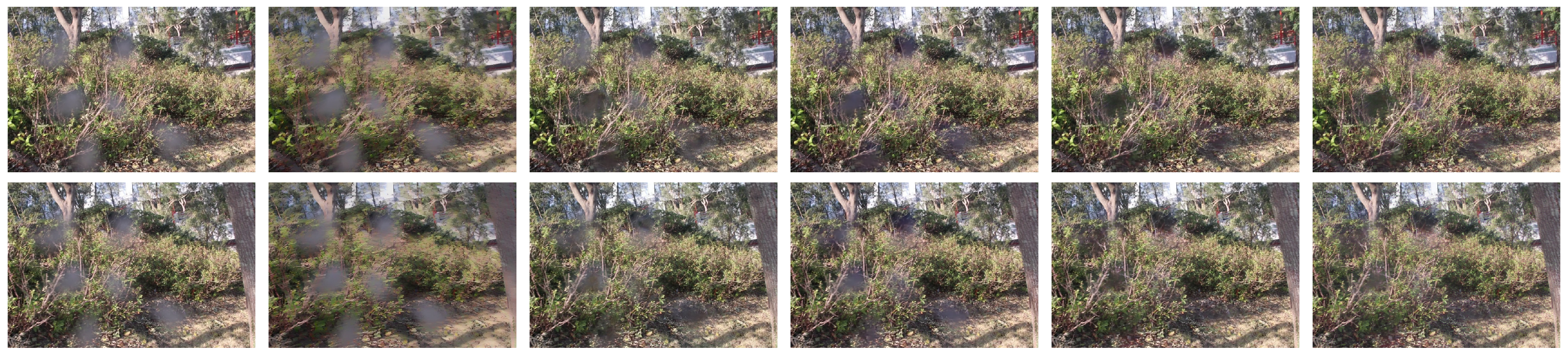}
	\begin{tabularx}{\linewidth}{YYYYYY}
        Input & FastDerain~\cite{jiang2018fastderain} & ObsRemoval~\cite{liu2020learning} & FastDVDnet~\cite{tassano2020fastdvdnet} & Ours (stage one) & Ours \\
    \end{tabularx}
    \vspace{-0.4cm}
	\caption[width=\textwidth]{Qualitative comparison with video-based restoration methods on the real video frames.}
	\vspace{-0.4cm}
	\label{fig:comparison_video}
\end{figure*}

\begin{table}[t]
\centering
\footnotesize
\begin{tabular}{@{}lcccc@{}}
	\toprule
    Method & Type & PSNR $\uparrow$ & SSIM $\uparrow$ & $E_{warp}$ $\downarrow$ \\
    \midrule
	PReNet~\cite{ren2019progressive}   & Single-image & 33.78 & 0.977 & 0.0049 \\
	AttGAN~\cite{qian2018attentive} & Single-image & 35.05 & \textbf{0.980} & 0.0047 \\
	FastDerain~\cite{jiang2018fastderain} & Video-based & 20.80 & 0.794 & 0.0075 \\
	ObsRemoval~\cite{liu2020learning}  & Video-based & 29.17 & 0.952 & 0.0052 \\
	FastDVDnet~\cite{tassano2020fastdvdnet} & Video-based & 31.95 & 0.936 & 0.0051 \\
	\cmidrule(lr){1-5}
	Ours (stage one)  & Video-based & \textbf{35.37} & \textbf{0.980} & 0.0045 \\
	Ours & Video-based & 34.98 & 0.979 & \textbf{0.0035} \\
	\bottomrule
\end{tabular}
\vspace{-0.2cm}
\caption{Quantitative comparison of our approach with other methods.}
\label{tab:tab_comparison}
\vspace{-0.2cm}
\end{table}

We validate the effectiveness of three main components in the first stage: attention detection, flow completion and spatial restoration. To keep all the other modules intact, we remove the attention detection module by giving attention maps with all zeros to the subsequent modules. For the removal of the flow completion, we directly use the degraded flow without any processing. In addition to using metrics such as PSNR and SSIM, we also use the warping error ($E_{warp}$) from~\cite{lai2018learning} to measure the temporal consistency of the results, \ie, we apply the method in~\cite{sundaram2010dense} to detect the occlusion regions and calculate the consistency between every two consecutive frames excluding these pixels. As shown in Table~\ref{tab:tab_single}, the best result is achieved in terms of PSNR, SSIM and $E_{warp}$ when the full model is used. Examples from the test set are shown in Figure~\ref{fig:ablation_image}. One can see that the proposed full model can generate results with fewer visible artifacts.

\vspace{-0.4cm}
\paragraph{Recurrent restoration.} Owing to the recurrent design of our network, our method is able to restore the frame progressively by iteratively utilizing the adjacent frames as shown in Figure~\ref{fig:res_recurrent}. In this example, four iterations of temporal fusion are used to obtain the output. The quantitative results in terms of PSNR and SSIM are shown in Figure~\ref{fig:res_frmnum}. In our test set, we take nine consecutive frames to restore the intermediate frame by providing the remaining eight frames progressively. Our method supports arbitrary number of frames during testing as opposed to~\cite{liu2020learning}.

\vspace{-0.4cm}

\paragraph{Running time.} We evaluate the inference time of our two-stage method on our test images at the resolution of $384 \times 384$. It takes approximately 1.358 seconds to produce one frame on a workstation with two Intel Xeon Gold 6244 CPUs and one Nvidia GeForce RTX-2080Ti GPU, where the first stage takes 0.987 seconds and the second stage takes 0.371 seconds. Note that the first stage will go through four iterations to refine a frame, while the second stage only needs one iteration to process each frame.

\subsection{Comparisons}

We compare our method with related techniques for single-image and video-based restoration on our test set. Five competitive methods with public source code are included, which are PReNet~\cite{ren2019progressive}, AttGAN~\cite{qian2018attentive}, FastDerain~\cite{jiang2018fastderain}, ObsRemoval~\cite{liu2020learning} and FastDVDnet~\cite{tassano2020fastdvdnet}. Among them, FastDerain~\cite{jiang2018fastderain} is an optimization-based method and others are learning-based approaches which are retrained on the same training set. These methods focus on different restoration tasks like adherent raindrop removal~\cite{qian2018attentive}, rain streak removal/deraining~\cite{ren2019progressive, jiang2018fastderain}, obstruction removal~\cite{liu2020learning}, and video denoising~\cite{tassano2020fastdvdnet}, which could be potentially applied to our task. As shown in Table~\ref{tab:tab_comparison},  our single-frame stage outperforms other methods in terms of PSNR and SSIM whereas the full model with the multi-frame processing achieves the lowest warping error. Figure~\ref{fig:comparison_single} and~\ref{fig:comparison_video} showcase the results on real scenes for qualitative comparison. Our method generalizes well to the real captured images and demonstrates more visually pleasing results without noticeable contaminant artifacts. Figure~\ref{fig:teaserfigure} shows that our method has the ability to remove various contaminants in the real world and produces high-quality results. We provide additional results in conjunction with the video outputs in the \emph{supplementary material}.
\section{Conclusion}

We present a novel framework that removes the contaminant artifact for moving cameras. We propose an attention detection module to localize the degraded regions and a flow completion module to recover the background motion for better alignment. Guided by the attention map and the restored flows, we recurrently fuse corresponding clean pixels to the current frame using the reference frames. Ultimately a multi-frame processing stage improves the temporal consistency. Experiments on both synthetic dataset and real scenes verify the effectiveness of each component and prove the quality advantage over prior approaches. We will make the synthetic dataset along with the source code publicly available and hopefully benefit the following works.

\newpage

{\footnotesize
\bibliographystyle{ieee_fullname}
\bibliography{egbib}
}

\end{document}